\documentclass[10pt,twocolumn,letterpaper]{article}
\usepackage[pagenumbers]{cvpr}
\def\confName{CVPR}
\def\confYear{2027}
\newcommand{\papertitle}{ROSE-Drive: Runtime Orchestration and System Evolution\\for Autonomous Driving}

\newcommand{\method}{ROSE-Drive\xspace}
\newcommand{\runOneDS}{97.71}
\newcommand{\runOneSR}{92.73}

\newcommand{\runTwoDS}{95.20}
\newcommand{\runTwoSR}{89.09}

\newcommand{\runThreeDS}{96.09}
\newcommand{\runThreeSR}{90.00}

\newcommand{\runFourDS}{97.66}
\newcommand{\runFourSR}{93.18}

\newcommand{\runFiveDS}{94.89}
\newcommand{\runFiveSR}{87.73}

\newcommand{\runSixDS}{94.27}
\newcommand{\runSixSR}{86.36}

\newcommand{\runNineIDDS}{89.0}
\newcommand{\runNineIDSR}{91.0}
\newcommand{\runNineIDHM}{90.0}
\newcommand{\runNineOODDS}{85.8}
\newcommand{\runNineOODSR}{87.0}
\newcommand{\runNineOODHM}{86.4}

\newcommand{\runTenIDDS}{79.6}
\newcommand{\runTenIDSR}{73.0}
\newcommand{\runTenIDHM}{76.2}
\newcommand{\runTenOODDS}{69.9}
\newcommand{\runTenOODSR}{55.0}
\newcommand{\runTenOODHM}{61.6}

\newcommand{\fTwoSimLingoDS}{87.3}
\newcommand{\fTwoSimLingoBaseDS}{89.2}
\newcommand{\fTwoSimLingoBaseSR}{95.0}
\newcommand{\fTwoSimLingoBaseHM}{92.0}
\newcommand{\fTwoSimLingoGenDS}{85.4}
\newcommand{\fTwoSimLingoGenSR}{89.0}
\newcommand{\fTwoSimLingoGenHM}{87.1}
\newcommand{\fTwoSimLingoAverageSR}{92.0}
\newcommand{\fTwoSimLingoAverageHM}{89.6}
\newcommand{\fTwoAutoMoTDS}{74.8}
\newcommand{\fTwoAutoMoTHarnessDS}{87.4}
\newcommand{\fTwoAutoMoTAverageSR}{64.0}
\newcommand{\fTwoAutoMoTAverageHM}{68.9}
\newcommand{\fTwoAutoMoTHarnessAverageSR}{89.0}
\newcommand{\fTwoAutoMoTHarnessAverageHM}{88.2}
\newcommand{\fTwoStageTwoAutoMoTDS}{81.3}
\newcommand{\fTwoStageTwoAutoMoTSR}{82.0}

\newcommand{\baseFourDS}{84.32}
\newcommand{\baseFourSR}{66.36}
\newcommand{\gainFourDS}{13.34}
\newcommand{\gainFourSR}{26.82}

\newcommand{\longestBaselineDS}{25}
\newcommand{\longestBaselineRC}{70}

\newcommand{\longestStageOneDS}{36}
\newcommand{\longestStageOneRC}{71}

\newcommand{\longestStageTwoDS}{44}
\newcommand{\longestStageTwoRC}{83}

\newcommand{\longestStageThreeDS}{52}
\newcommand{\longestStageThreeRC}{91}

\newcommand{\latencyBaseMedian}{171.3}
\newcommand{\latencyBasePNinety}{180}
\newcommand{\latencyHarnessPNinety}{340}
\newcommand{\latencySimLingoBasePNinety}{660}
\newcommand{\latencySimLingoHarnessPNinety}{840}

\newcommand{\fTwoBridgeHarnessBaseDS}{91.6}

\newcommand{\fTwoBridgeHarnessBaseSR}{99.0}
\newcommand{\fTwoBridgeHarnessBaseHM}{95.1}
\newcommand{\fTwoBridgeHarnessGenDS}{88.9}

\newcommand{\fTwoBridgeHarnessGenSR}{94.0}
\newcommand{\fTwoBridgeHarnessGenHM}{91.4}
\newcommand{\fTwoBridgeHarnessAverageDS}{90.3}

\newcommand{\fTwoBridgeHarnessAverageSR}{96.5}
\newcommand{\fTwoBridgeHarnessAverageHM}{93.3}

\newcommand{\latencyBaseMean}{138.9}
\newcommand{\latencyHarnessMedian}{107.2}

\newcommand{\latencyHarnessMean}{155.2}

\newcommand{\longestPolicySkipShare}{71.5}

\newcommand{\longestPolicyCallShare}{28.5}

\newcommand{\latencySimLingoBaseMedian}{521.3}
\newcommand{\latencySimLingoBaseMean}{539.0}
\newcommand{\latencySimLingoHarnessMedian}{103.2}
\newcommand{\latencySimLingoHarnessMean}{280.9}
\newcommand{\longestSimLingoSkipShare}{71.3}
\newcommand{\longestSimLingoCallShare}{28.7}

\newcommand{\longestSimLingoBaselineDS}{22}
\newcommand{\longestSimLingoBaselineRC}{70}
\newcommand{\longestSimLingoStageOneDS}{36}
\newcommand{\longestSimLingoStageOneRC}{73}
\newcommand{\longestSimLingoStageTwoDS}{39}
\newcommand{\longestSimLingoStageTwoRC}{80}
\newcommand{\longestSimLingoStageThreeDS}{47}
\newcommand{\longestSimLingoStageThreeRC}{84}
\newcommand{\baseOneDS}{89.68}
\newcommand{\baseOneSR}{75.45}
\newcommand{\gainOneDS}{8.03}
\newcommand{\gainOneSR}{17.28}

\usepackage[T1]{fontenc}
\usepackage{microtype}
\usepackage{mathtools,bm,multirow,array,tabularx}
\usepackage{pifont}
\usepackage{colortbl}
\definecolor{releasedreferencegray}{gray}{0.93}
\definecolor{releasedreferenceink}{gray}{0.40}
\usepackage{placeins}
\usepackage{algorithm,algpseudocode}
\definecolor{cvprblue}{rgb}{0.21,0.49,0.74}
\ifdefined\ReviewMode
  \usepackage[pagebackref,breaklinks,colorlinks,allcolors=cvprblue]{hyperref}
\else
  \usepackage[breaklinks,colorlinks,allcolors=cvprblue]{hyperref}
\fi
\newcommand{\E}{\mathcal E}

\newcommand{\Nav}{\mathsf{nav}}
\newcommand{\Lat}{\mathsf{lat}}
\newcommand{\Lon}{\mathsf{lon}}

\newcommand{\valid}{\operatorname{valid}}
\newcommand{\bind}{\operatorname{bind}}

\newcolumntype{Y}{>{\raggedright\arraybackslash}X}
\graphicspath{{figures/}}
\hypersetup{pdftitle={ROSE-Drive: Runtime Orchestration and System Evolution for Autonomous Driving},pdfauthor={},pdfsubject={System-level orchestration and offline system evolution with frozen driving models}}
\newcommand{\gain}[1]{\rlap{\,\textsuperscript{\scriptsize\textcolor{cvprblue}{+#1}}}}

\usepackage{stfloats}
\newcommand{\arxivauthors}{%
Qihang Huang\textsuperscript{1}\quad
Zhidong Wang\textsuperscript{1}\quad
Zirui Li\textsuperscript{1}\quad
YuJie Yan\textsuperscript{1}\quad
Jingsong Liang\textsuperscript{1}\\
Shiyu Fang\textsuperscript{2}\quad
Collister Chua\textsuperscript{1}\quad
Zhan Chen\textsuperscript{1}\quad
Chen Lv\textsuperscript{1*}\\[0.3em]
\textsuperscript{1}Nanyang Technological University\quad
\textsuperscript{2}Tongji University}
\newcommand{\arxivauthornames}{Qihang Huang, Zhidong Wang, Zirui Li, YuJie Yan, Jingsong Liang, Shiyu Fang, Collister Chua, Zhan Chen, Chen Lv}

\title{\papertitle}
\author{\arxivauthors}
\hypersetup{pdfauthor={\arxivauthornames}}
\begin{document}
\twocolumn[{
\renewcommand\twocolumn[1][]{#1}
\maketitle
\begin{minipage}{\textwidth}
\centering
\includegraphics[width=\textwidth]{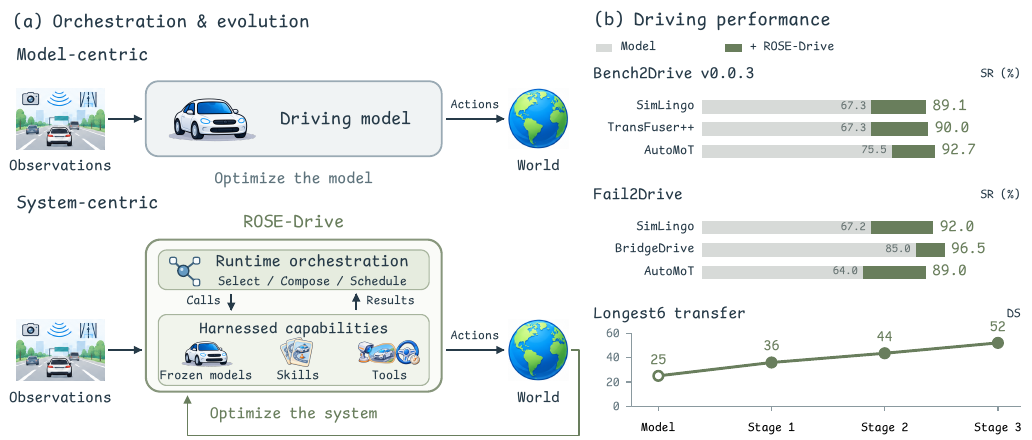}
\captionof{figure}{\textbf{System orchestration and accumulated gains.} (a) \method selects, combines, and schedules models, skills, and tools. System evolution refines skill implementations and coordination without retraining the driving model or VLM. (b) Success-rate gains on Bench2Drive v0.0.3 and Fail2Drive, and stage-wise driving scores on Longest6 with AutoMoT fixed.}
\label{fig:motivation}
\end{minipage}
\vspace{0.5em}

}]
\begingroup
\renewcommand{\thefootnote}{*}
\footnotetext[1]{Corresponding author.}
\endgroup
\begin{abstract}
End-to-end driving models have made substantial progress, yet reliable closed-loop driving in complex traffic remains challenging. Beyond improving individual models, system-level orchestration offers a complementary approach by coordinating driving capabilities and computation. We introduce \method (Runtime Orchestration and System Evolution), a model-agnostic framework that jointly orchestrates driving execution and computation. It coordinates frozen driving models with driving skills, perception tools, and feedback controllers. A distilled vision--language model (VLM) proposes driving tasks and scene relations. Shared physical state and persistent tasks support capability composition, task continuity, and adaptive scheduling. The orchestrator combines model and skill outputs across control channels, maintains task responsibility through handover, and uses execution feedback to select controllers, reuse valid motion references, and schedule model inference. Offline revisions use execution records to refine skills and their coordination through a stable task interface. Successive system stages inherit these revisions while model, perception, and VLM weights remain fixed. Evaluations with four driving models on Bench2Drive v0.0.3, Bench2Drive v0.0.4, and Fail2Drive show accumulated capability and transfer before target-benchmark revision. On held-out Longest6, driving score (DS) rises from 25 to 52 for AutoMoT and from 22 to 47 for SimLingo across system stages, while both systems skip new model inference on over 71\% of control steps.
\end{abstract}

\section{Introduction}
\label{sec:intro}
End-to-end driving policies have made substantial progress across diverse traffic scenarios~\cite{uniad,diffusiondrive,simlingo,automot,interact}. Beyond improving individual policies, a complementary direction is to coordinate their use with driving skills, perception tools, and feedback controllers at the system level. A runtime harness exposes these capabilities through common interfaces. System-level orchestration determines how their contributions are combined and when new computation is needed.

The central difficulty is coordinating these choices throughout an ongoing maneuver. A skill can regulate speed while retaining a compatible model path. When the model supplies both motion channels, persistent tracking error can trigger a controller change. A navigation change can prompt a new model call. The maneuver may also need to continue after the latest scene assessment no longer requests it. These decisions require task state that persists across changing observations and capability outputs.

We introduce \method, a model-agnostic framework that jointly orchestrates driving execution and computation (Fig.~\ref{fig:motivation}). A distilled vision--language model (VLM) proposes driving tasks and scene relations. The orchestrator binds these proposals to observed objects and road regions to form task requests. Shared physical state and persistent tasks guide capability composition, task continuity, and adaptive scheduling. These mechanisms determine which outputs to combine, how to maintain responsibility through a maneuver, and when computation is needed.

A shared task interface also supports cumulative improvement. Skills address physical requirements that recur across benchmarks, including traffic-interaction timing and obstacle clearance. Current geometry and motion determine each task's constraints, allowing the same implementation to handle different objects and layouts. The VLM's relation vocabulary and skill families stay fixed as execution records guide offline improvements to skill implementations and orchestration rules. Successive system versions, called stages, inherit these improvements.

We study four frozen models across Bench2Drive v0.0.3, v0.0.4, and Fail2Drive, including transfer before target-specific revision. Longest6 is reserved for evaluating frozen systems throughout this process. AutoMoT's driving score (DS) progresses from 25 alone to 36, 44, and 52 across system stages, while SimLingo progresses from 22 to 36, 39, and 47. Both systems skip model inference on over 71\% of Longest6 control steps. Our contributions are:
\begin{itemize}
\item \textbf{Model-Agnostic Runtime Framework.} We introduce a unified architecture that connects frozen driving models, skills, perception tools, and feedback controllers through shared physical state and a common task interface. Model adapters preserve each model's inputs and outputs while making it available to the same runtime.
\item \textbf{Coordinated Execution and Scheduling.} The orchestrator combines model and skill outputs across control channels while persistent tasks maintain responsibility through handover. The same task state and execution feedback guide controller selection, motion-reference reuse, and model inference, coupling capability composition, task continuity, and adaptive scheduling.
\item \textbf{System Evolution and Transfer.} A stable task interface supports offline refinement of shared skills and orchestration rules with model, perception, and VLM weights fixed. Successive versions inherit these changes and transfer across benchmarks and driving models.
\end{itemize}

\section{Related Work}
\label{sec:related}
\paragraph{Driving Models and Hybrid Planning.}
End-to-end policies integrate perception and planning~\cite{tcp,uniad,drivetransformer}. Generative and vision--language models expand their driving behavior~\cite{diffusiondrive,occwam,bridgedrive,lmdrive,simlingo,automot}, while temporal and closed-loop learning further improve these policies~\cite{momad,craft}. Hybrid systems combine learned and structured components. PDM-Hybrid combines rule-based planning with learned long-term waypoint corrections~\cite{pdmhybrid}, and DriveVLM-Dual couples semantic reasoning with planning~\cite{drivevlm}. AsyncDriver and AutoMoT decouple reasoning from motion prediction~\cite{asyncdriver,automot}. \method builds on frozen driving models and coordinates their use with skills and controllers during execution.

\paragraph{Driving Agents.}
Driving agents use language reasoning and external capabilities to guide driving decisions. Agent-Driver, DiLu, and LeapAD incorporate tools and experience into this process~\cite{agentdriver,dilu,leapad}. DrivingAgent designs and validates modules and trains an online scheduling agent~\cite{drivingagent}. The concurrent AutoAgent0 runtime verifies nominal trajectories against refined scene context and generates recovery proposals when none is admitted~\cite{autoagentzero}. Argus supplies temporary mitigation trajectories~\cite{argus}. These two runtimes intervene through motion proposals. \method coordinates model and skill contributions within a maneuver. The model can supply steering while a skill supplies speed, and the task retains responsibility through handover. Our execution comparisons examine this cooperation, task continuity, and adaptive scheduling (Sec.~\ref{sec:component-comparisons}).

\paragraph{Embodied Agent Harnesses.}
Robot execution has long used behavior trees to organize task control flow~\cite{behaviortrees} and runtime-assurance architectures to supervise controller switching~\cite{simplex}. Agent harnesses connect semantic reasoning to physical execution through shared context, callable capabilities, and execution contracts~\cite{thea,phyagentos}. Harness VLA exposes a frozen VLA as a contact-rich primitive alongside analytic primitives~\cite{harnessvla}, while RoboHarness uses execution memory to route subtasks among heterogeneous policies and prepare compatible handoffs~\cite{roboharness}. DynaHarness, HarnessPAI, RoboFoundry, and Code-Only-as-Policy (COAP) revise reusable execution capabilities or programs~\cite{dynaharness,harnesspai,robofoundry,coap}. \method coordinates simultaneous contributions on navigation, lateral, and longitudinal channels, using persistent task state to select controllers and schedule model inference (Appendices~A and~C). Its fixed relation vocabulary and shared skill contracts allow improved implementations and coordination rules to carry across system versions.

\section{ROSE-Drive}
\label{sec:method}
\begin{figure*}[t]
\centering
\includegraphics[width=\textwidth,height=302bp,keepaspectratio]{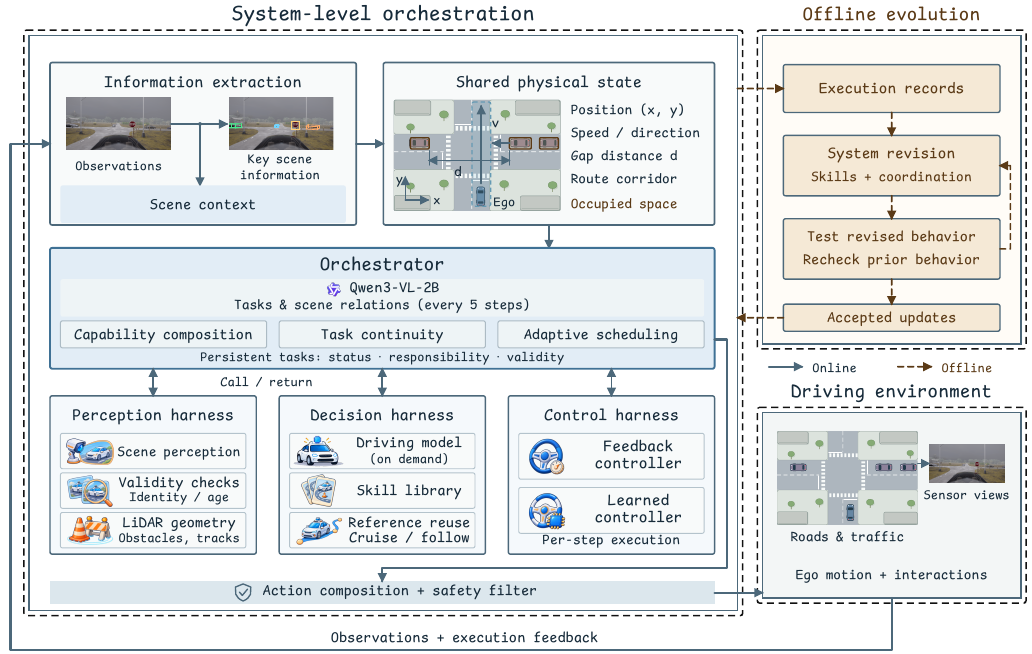}
\caption{\textbf{\method system architecture.} Capability composition selects and combines outputs. Task continuity maintains responsibility through execution and handover. Adaptive scheduling determines when to run the driving model and reuse motion references. The VLM proposes driving tasks and scene relations every five control steps, while the orchestrator updates task state at every step. The three harnesses expose perception tools, models, skills, and controllers. The safety filter checks the resulting commands. Solid and brown dashed arrows denote online execution and offline revision.}
\label{fig:architecture}
\end{figure*}

\method coordinates driving capabilities through shared physical state and persistent tasks (Fig.~\ref{fig:architecture}). Shared physical state $\E_t$ records tracked objects, geometry, motion, observation times, and validity. Perception, decision, and control harnesses expose capabilities to the orchestrator, while skill contracts define when and how each skill participates in a task.
\subsection{Task State and Model Interface}
\label{sec:formulation}
Capability composition uses navigation, lateral, and longitudinal channels $\mathcal W=\{\Nav,\Lat,\Lon\}$. The navigation channel adjusts model inputs. Lateral and longitudinal outputs are native commands or motion references (target paths or speeds for tracking). Adapters preserve each model's observations and output conventions.

A task maintains a driving requirement as the scene changes, while a skill supplies its execution logic. Task $\tau_i=(b_i,\phi_i,W_i)$ stores scene bindings $b_i$, status $\phi_i$, and required channels $W_i\subseteq\mathcal W$. A binding links the task to observed entities or road regions.

Channel ownership $\alpha_t(c)$ identifies the task responsible for output selection and recovery, with $0$ denoting no owner. Output source $s_t(c)$ identifies whose output is applied, with $0$ denoting the driving model and $i$ the skill instance serving $\tau_i$. Each channel uses one source, and unowned channels default to the model. A task can retain ownership while using compatible model output. Controller $k_t$ executes the selected output, and $a_t$ requests model inference. Validity checks test scene bindings and motion references against current observations.

Algorithm~\ref{alg:system-cycle} shows how execution feedback updates tasks, determines model calls, and guides channel selection and handover within each control step.
\begin{algorithm}[t]
\caption{Runtime orchestration loop}
\label{alg:system-cycle}
\small
\begin{algorithmic}[1]
\algtext*{EndIf}
\algtext*{EndFor}
\Require Frozen model and VLM, skill contracts
\Statex Persistent state: tasks, motion references, applied commands
\For{each control step $t$}
\State Update shared physical state, task progress, and feedback
\For{each task already in a model trial}
\State Review applied commands and resulting motion
\State \textbf{if trial accepted:} release task ownership
\State \textbf{else if rejected:} resume skill execution or recovery
\State \textbf{else} retain ownership and continue the trial
\EndFor
\State \textbf{if new VLM proposals:} bind and update task requests
\State Admit tasks with available channels (Eq.~\ref{eq:admit})
\State \textbf{if a task owns navigation:} apply valid updates
\State Check motion references and skill plans
\State Identify model outputs needed for execution or handover
\If{new model output is required (Eq.~\ref{eq:activation})}
\State Run the model and refresh cached references
\Else
\State Retain valid references, skill plans, or a required hold
\EndIf
\State Generate skill outputs for the updated task state
\For{each Active task requesting handover}
\State \textbf{if completion or readiness fails:} retain ownership
\State \textbf{else if trial required:} start trial with ownership
\State \textbf{else} return channels under the skill contract
\EndFor
\State Select compatible model and skill outputs per channel
\State Select the tracking controller using execution feedback
\State Track references and forward selected native commands
\State Apply safety filter and execute the composed command
\State Retain task state and the command actually applied
\EndFor
\end{algorithmic}
\end{algorithm}

\subsection{VLM Proposals and Shared Physical State}
\label{sec:student}
The perception harness maintains $\E_t$ by associating detections, tracks, and occupancy. It checks whether observations are current and sufficient for the task. Shared scene bindings keep task selection, motion generation, and execution checks tied to the same interaction.

Geometric rules generate candidate scene relations $\mathcal C_t$. A distilled Qwen3-VL-2B evaluates them to propose driving tasks and scene relations $\mathcal R_t$. The orchestrator binds these proposals to observed objects and road regions, forms task requests, and checks skill applicability and channel availability before admission. Ongoing tasks continue according to execution feedback even without new VLM proposals.

The VLM is trained open-loop using PDM-Lite expert contexts and teacher-reviewed relation labels, then frozen throughout system evolution and evaluation. VLM inference is scheduled asynchronously every five control steps, while the driving model is queried on demand. Appendix~A describes training and scheduling. Appendix~B details the shared physical state.

\subsection{Skill Library}
\label{sec:skill-library}
Four skill families address recurring physical requirements (Table~\ref{tab:skill-library}). Observed objects and road regions determine each task's geometry and motion constraints. Five VLM relation families identify these requirements, with emergency yielding and traffic rules handled by the same rule-compliance family.
\begin{table}[t]
\centering
\caption{\textbf{Skill library.} Five VLM relation families map to four reusable skill families.}
\label{tab:skill-library}
\small\setlength{\tabcolsep}{3pt}
\renewcommand{\arraystretch}{1.08}
\begin{tabularx}{\linewidth}{@{}l>{\raggedright\arraybackslash}Xl@{}}
\toprule
Relation & Requirement & Skill family\\
\midrule
Traffic interaction & Timing and speed & Traffic interaction\\
\addlinespace[2pt]
Obstacle passage & Clearance, timing,\newline and lane~return & Obstacle passage\\
\addlinespace[2pt]
Emergency yielding & Emergency priority & Rule compliance\\
\addlinespace[2pt]
Traffic rule & Signal / STOP rules & Rule compliance\\
\addlinespace[2pt]
Route guidance & Navigation targets & Route guidance\\
\bottomrule
\end{tabularx}
\end{table}

Traffic interaction plans departure timing and speed under predicted traffic occupancy along an available path, including a model path. Crossings and merges share this space--time planner, with geometry, traffic motion, and conflict type determining its constraints. Obstacle passage plans a path around obstructions and back to the lane, with timing constrained by tracked traffic. Obstacle extent and free road boundaries determine clearance. Different obstructions change these constraints while reusing the same skill implementation. Rule compliance maintains traffic obligations and task progress across observations (Appendix~C).

The skill instance serving task $\tau_i$ exposes a contract
\begin{equation}
K_i=(\Omega_i,W_i,\kappa_i,d_i,h_i),
\label{eq:skill-contract}
\end{equation}
specifying applicability, required channels, output generation, completion, and handover readiness. Route guidance adjusts navigation inputs so the model can generate the required motion. The common contract lets each skill retain its own solver (Appendix~C). Revisions can refine these conditions and implementations while preserving the VLM's relation vocabulary and the contract structure.

\subsection{Capability Composition}
\label{sec:interface}\label{sec:authority}
A task receives its required channels when its scene bindings are valid, the skill is applicable, and the channels are available:
\begin{equation}
\begin{aligned}
\operatorname{admit}(i)={}&\valid(b_i)\land\Omega_i(\E_t,b_i)\\
&{}\land\!\bigwedge_{c\in W_i}[\alpha_t(c)\in\{0,i\}].
\end{aligned}
\label{eq:admit}
\end{equation}
Channels are allocated together and are not taken from their current owner. Tasks using disjoint channel sets can coexist. An active task can change its required set as execution progresses.

A traffic-interaction task can retain a compatible model path while its skill regulates speed. Selection respects current task ownership, output validity, and compatibility across channels:
\begin{equation}
\tilde z_t^c=z_t^{s_t(c),c},\qquad
u_t=\mathcal G\!\left(\operatorname{Exec}_{k_t}(\tilde z_t,x_t),\E_t\right),
\label{eq:compose}
\end{equation}
where $c\in\{\Lat,\Lon\}$, $z_t^{s,c}$ is a reference or native command, and $\tilde z_t$ collects the selected outputs. $\operatorname{Exec}_{k_t}$ tracks selected references using ego state $x_t$ and forwards native commands on their selected channels. Controller selection respects channel ownership and reference validity. During model-reference tracking, persistent lateral, heading, or speed error can trigger a switch from a feedback controller (PID) to a frozen learned controller. Sustained low tracking error restores feedback control. Safety filter $\mathcal G$ checks the composed command. The command actually applied and subsequent observations supply execution feedback (Appendix~C). Table~\ref{tab:execution-example} illustrates capability composition and reference reuse during a merge.

\subsection{Task Continuity and Handover}
\label{sec:transition}
Task completion $d_i$ and handover readiness $h_i$ serve different purposes. Completion checks whether the task has fulfilled its requirement. Readiness checks whether the model can continue from the resulting physical state using valid outputs. A task is \textbf{Active} during waiting, execution, and handover checks, \textbf{Monitor} during a model trial, and \textbf{Idle} after termination. A task can return channels before terminating. A pending request has not acquired channels.

For traffic interaction, handover readiness starts a monitored model trial. The output source switches to the model while the task retains ownership:
\begin{equation}
s_t(c)=0,\qquad \alpha_t(c)=i,\qquad c\in W_i.
\label{eq:monitored-trial}
\end{equation}
Trials that began in earlier steps are reviewed using the commands actually applied and the resulting motion. Forward motion or waiting justified by current traffic permits release of task ownership. The task retains responsibility during the trial and can resume control if the model cannot continue. This review precedes new output generation. Other skill contracts can return channels without a trial. Figure~\ref{fig:cases} illustrates task continuity and handover. Appendix~D details the handover conditions.
\begin{table}[t]
\centering
\caption{\textbf{Capability composition within a persistent task.} During a merge, model steering and skill speed coexist while the task remains active. After release, a cached model path supports feedback control without new model inference.}
\label{tab:execution-example}
\small\setlength{\tabcolsep}{3pt}
\begin{tabularx}{\linewidth}{@{}lc>{\raggedright\arraybackslash}Xc@{}}
\toprule
Time (s) & Task status & Lateral / longitudinal & Model call\\
\midrule
4.45 & Active & Model steering / skill speed & Yes\\
10.05 & Active & Model steering / skill speed & Yes\\
14.05 & Idle & Cached path / speed tracking & No\\
\bottomrule
\end{tabularx}
\end{table}

\subsection{Adaptive Scheduling}
\label{sec:activation}
Task state and execution feedback determine model inference. Between calls, controllers track valid references and adjust cruising or following from current observations. Inference can be skipped when a required hold ($\eta_t^{\rm hold}$), valid model reference ($\eta_t^{\rm reuse}$), or executable skill plan ($\eta_t^{\rm plan}$) supplies the required outputs. With $\eta_t^{\rm refresh}$ indicating a required model call,
\begin{equation}
a_t=\eta_t^{\rm refresh}\lor\neg(\eta_t^{\rm hold}\lor\eta_t^{\rm reuse}\lor\eta_t^{\rm plan}).
\label{eq:activation}
\end{equation}
$a_t=1$ runs the model and updates its cached references. Navigation changes, departure after a traffic-rule stop, invalid motion references, and safety-filter requests set $\eta_t^{\rm refresh}$. Pending handover also sets it when fresh model output is required. The orchestrator then checks handover readiness using valid outputs and retains task responsibility if the conditions are unmet. Tracking and safety checks continue between model calls. Appendix~A gives the reference validity checks.

\subsection{System Evolution}
\label{sec:evolution}
The relation vocabulary, skill contracts, and control channels form a stable task interface. Across stages, the VLM proposes the same types of driving tasks and scene relations, while execution records guide offline revisions to skill implementations and orchestration rules. Crossings and merges, for example, share a traffic-interaction planner whose constraints are instantiated from each task's current geometry and motion. Improvements to this shared implementation can therefore be reused by both, and later stages inherit them through the same interface.

Revisions also refine the construction of shared physical state, controller selection, reference reuse, and model inference. Coding agents analyze records, implement revisions, and check execution. Model, perception, and VLM weights remain fixed across stages. Frozen-stage evaluations assess their transfer across benchmarks and driving models (Appendix~E).

\section{Experiments}
\label{sec:experiments}
We evaluate driving performance across models and benchmarks, test whether successive system improvements transfer, and examine the contributions of capability composition, task continuity, and adaptive scheduling.
\begin{figure*}[t]
\centering
\includegraphics[width=\textwidth]{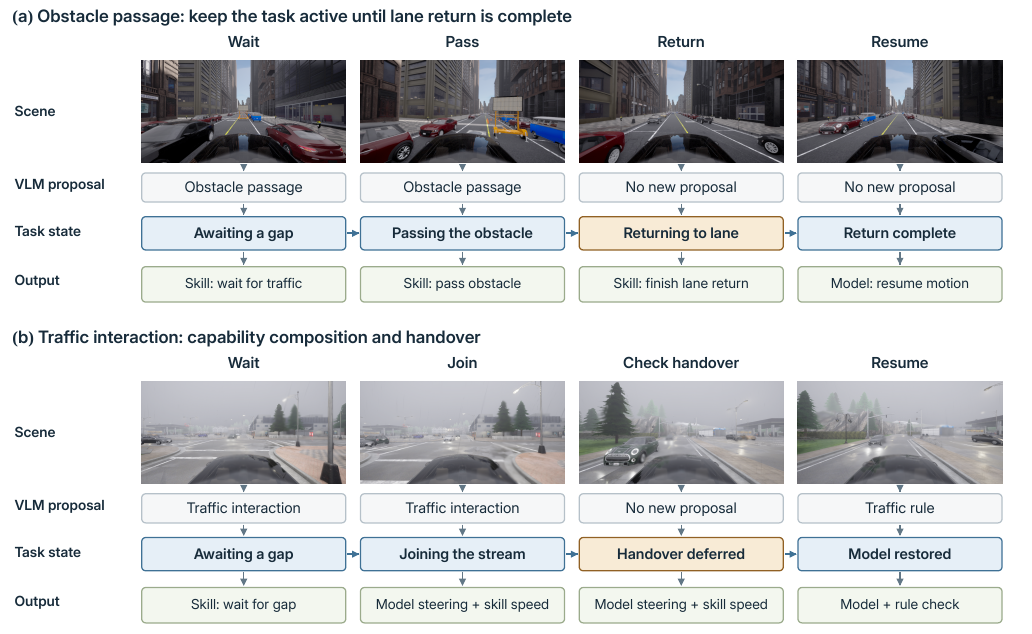}
\caption{\textbf{Capability composition and task continuity in execution.} Vertical arrows link observations, VLM proposals, task state, and applied outputs. Horizontal arrows carry task state across updates. (a) The orchestrator completes lane return after the obstacle-passage proposal disappears. (b) Model steering and skill speed remain combined while handover is deferred. After handover readiness checks and a monitored model trial, the task releases ownership. Ochre cards highlight active tasks without new VLM proposals.}
\label{fig:cases}
\end{figure*}
\suppressfloats[t]

\begin{table}[t]
\centering
\caption{\textbf{Model and system performance on Bench2Drive v0.0.3.} AutoMoT uses our route evaluations. SimLingo and TransFuser++ scores follow the Bench2Drive leaderboard~\cite{b2dexpertrefs}. Blue superscripts show gains over the model alone. $\dagger$ marks privileged data-collection experts using ground-truth scene state~\cite{b2dexpertrefs}.}
\label{tab:b2d003}\label{tab:b2d}\label{tab:b2dtransfer}
\small\setlength{\tabcolsep}{4pt}
\begin{tabularx}{\linewidth}{@{}Xr@{\hspace{22pt}}r@{\hspace{20pt}}}
\toprule Model / system & DS $\uparrow$ & SR (\%) $\uparrow$\\
\midrule
SimLingo & 85.07 & 67.27\\
\quad + \method & \runTwoDS\gain{10.13} & \runTwoSR\gain{21.82}\\
\midrule
TransFuser++ & 84.21 & 67.27\\
\quad + \method & \runThreeDS\gain{11.88} & \runThreeSR\gain{22.73}\\
\midrule
AutoMoT & \baseOneDS & \baseOneSR\\
\quad + \method & \textbf{\runOneDS}\gain{\gainOneDS} & \textbf{\runOneSR}\gain{\gainOneSR}\\
\midrule
\rowcolor{releasedreferencegray}[\tabcolsep][20pt]
\textcolor{releasedreferenceink}{PDM-Lite$^\dagger$} & \textcolor{releasedreferenceink}{97.02} & \textcolor{releasedreferenceink}{92.27}\\
\rowcolor{releasedreferencegray}[\tabcolsep][20pt]
\textcolor{releasedreferenceink}{Think2Drive$^\dagger$} & \textcolor{releasedreferenceink}{91.85} & \textcolor{releasedreferenceink}{85.41}\\
\bottomrule
\end{tabularx}
\end{table}

\begin{table}[t]
\centering
\caption{\textbf{Model and system performance on Bench2Drive v0.0.4.} SimLingo and the LEAD expert follow the release~\cite{b2d004}. AutoMoT and TransFuser++ are our evaluations. Blue superscripts show gains over the model alone. $\dagger$ marks a privileged expert using ground-truth scene state~\cite{lead}.}
\label{tab:b2d004}
\small\setlength{\tabcolsep}{4pt}
\begin{tabularx}{\linewidth}{@{}Xr@{\hspace{22pt}}r@{\hspace{20pt}}}
\toprule Model / system & DS $\uparrow$ & SR (\%) $\uparrow$\\
\midrule
SimLingo & 86.55 & 70.45\\
\quad + \method & \runFiveDS\gain{8.34} & \runFiveSR\gain{17.28}\\
\midrule
TransFuser++ & 86.52 & 70.00\\
\quad + \method & \runSixDS\gain{7.75} & \runSixSR\gain{16.36}\\
\midrule
AutoMoT & \baseFourDS & \baseFourSR\\
\quad + \method & \textbf{\runFourDS}\gain{\gainFourDS} & \textbf{\runFourSR}\gain{\gainFourSR}\\
\midrule
\rowcolor{releasedreferencegray}[\tabcolsep][20pt]
\textcolor{releasedreferenceink}{LEAD expert$^\dagger$} & \textcolor{releasedreferenceink}{95.59} & \textcolor{releasedreferenceink}{87.28}\\
\bottomrule
\end{tabularx}
\end{table}

\subsection{Experimental Setup}
Bench2Drive v0.0.3~\cite{bench2drive} and v0.0.4~\cite{b2d004} each contain 220 routes across the same 44 interaction types. Version v0.0.4 introduces a new route set. Fail2Drive adds 17 long-tail types and 30 assets through 100 Base--Gen route pairs testing appearance, layout, behavior, and robustness shifts~\cite{fail2drive}. Longest6 contains 36 routes~\cite{longest6} reserved for evaluating frozen systems. We report driving score (DS), success rate (SR), their harmonic mean (HM) on Fail2Drive, and route completion (RC) on Longest6.

We use one execution architecture across all reported benchmarks and develop successive system stages with AutoMoT. Bench2Drive v0.0.3 yields Stage~1, v0.0.4 Stage~2, and Fail2Drive Stage~3. Stage~2 also includes closed-loop runs on CARLA training routes. Stages inherit revised skills and coordination rules through the same task interface (Appendix~E). Driving-model and perception weights remain fixed, and the VLM is frozen after open-loop training. SimLingo~\cite{simlingo}, TransFuser++~\cite{hiddenmodels,hiddendata}, and BridgeDrive~\cite{bridgedrive} connect through model adapters.

Development combines corner cases that challenge the driving model or current system with regression routes used to check that model performance is preserved. Routes excluded from development form the held-out set (Appendix~E). Benchmark tables aggregate both sets. Transfer tests freeze Stage~1 before v0.0.4 revision and Stage~2 before Fail2Drive revision. Appendix~G details baseline sources.

AutoMoT, TransFuser++, and BridgeDrive already use TransFuser-family camera--LiDAR perception with detection and BEV-semantic heads. BridgeDrive uses the LEAD-trained TFv6 variant~\cite{bridgedrive,lead}. Shared physical state is built from frozen PDM-Lite-trained TransFuser++ perception, reusing AutoMoT's and TransFuser++'s existing weights. SimLingo retains camera-only model inputs and additionally receives this camera--LiDAR module at system level. Appendix~A details the configurations. CARLA~\cite{carla} runs synchronously.

\subsection{Closed-Loop Driving Performance}
\method reaches \runOneDS{} DS on Bench2Drive v0.0.3 and \runFourDS{} on v0.0.4 with AutoMoT (Tables~\ref{tab:b2d003} and~\ref{tab:b2d004}). Each benchmark uses one system across all 44 interaction types, with tasks selected from current observations through the shared interface.

\paragraph{Reuse across models.}
Using the same runtime through model adapters, SimLingo and TransFuser++ exceed 94 DS on both Bench2Drive versions. Gains range from 7.75 to \gainFourDS{} DS across the three models.

\paragraph{Fail2Drive.}
Stage~3 extends the inherited system to Fail2Drive's long-tail conditions, reaching \fTwoAutoMoTHarnessDS{} DS with AutoMoT, \fTwoSimLingoDS{} with SimLingo, and \fTwoBridgeHarnessAverageDS{} with BridgeDrive (Table~\ref{tab:f2d-local}). Appendix~G reports Base/Gen results and baseline sources.
\begin{table}[t]
\centering
\caption{\textbf{Model and system performance on Fail2Drive.} DS, SR (\%), and HM are averaged over splits. Blue superscripts show gains over the model alone. Appendix~G gives split results and baseline sources.}
\label{tab:f2d-local}
\small\setlength{\tabcolsep}{4pt}
\begin{tabularx}{\linewidth}{@{}Xr@{\hspace{21pt}}r@{\hspace{21pt}}r@{\hspace{18pt}}}
\toprule Model / system & DS $\uparrow$ & SR $\uparrow$ & HM $\uparrow$\\
\midrule
AutoMoT & \fTwoAutoMoTDS & \fTwoAutoMoTAverageSR & \fTwoAutoMoTAverageHM\\
\quad + \method & \fTwoAutoMoTHarnessDS\gain{12.6} & \fTwoAutoMoTHarnessAverageSR\gain{25.0} & \fTwoAutoMoTHarnessAverageHM\gain{19.3}\\
\midrule
SimLingo & 77.2 & 67.2 & 71.6\\
\quad + \method & \fTwoSimLingoDS\gain{10.1} & \fTwoSimLingoAverageSR\gain{24.8} & \fTwoSimLingoAverageHM\gain{18.0}\\
\midrule
BridgeDrive & 86.8 & 85.0 & 85.8\\
\quad + \method & \textbf{\fTwoBridgeHarnessAverageDS}\gain{3.5} & \textbf{\fTwoBridgeHarnessAverageSR}\gain{11.5} & \textbf{\fTwoBridgeHarnessAverageHM}\gain{7.5}\\
\bottomrule
\end{tabularx}
\end{table}

On held-out routes, where AutoMoT alone already reaches 97.63 DS on v0.0.3, \method maintains 97.91 and raises DS from 87.16 to 98.43 on v0.0.4 and from 79.46 to 91.75 on Fail2Drive (Appendix~E).

\subsection{System Evolution and Transfer}
\label{sec:stage-transfer}
Before any v0.0.4 revision, Stage~1 transfers the v0.0.3 system to the new route set. Across all 220 routes, AutoMoT DS rises from 84.3 to 95.1 and SimLingo from 86.6 to 92.7 (Table~\ref{tab:cross-benchmark-transfer}). This tests reuse across new layouts within the same 44 interaction types. Stage~2 transfers the expanded system to Fail2Drive, raising AutoMoT from \fTwoAutoMoTDS{} to \fTwoStageTwoAutoMoTDS{} DS and from \fTwoAutoMoTAverageSR\% to \fTwoStageTwoAutoMoTSR\% SR. SimLingo's SR rises from 67.2\% to 80.0\%. With SimLingo, the system constructed using AutoMoT also transfers across models through an adapter.
\begin{table}[t]
\centering
\caption{\textbf{Transfer of frozen system stages.}
Stage~1 transfers to all 220 Bench2Drive v0.0.4 routes and Stage~2 to Fail2Drive.
Model baselines follow Tables~\ref{tab:b2d004} and~\ref{tab:f2d-local}.}
\label{tab:cross-benchmark-transfer}
\small\setlength{\tabcolsep}{3pt}
\begin{tabularx}{\linewidth}{@{}Xrrrr@{}}
\toprule
& \multicolumn{2}{c}{DS $\uparrow$} & \multicolumn{2}{c}{SR (\%) $\uparrow$}\\
\cmidrule(lr){2-3}\cmidrule(l){4-5}
Model & Alone & + \method & Alone & + \method\\
\midrule
\multicolumn{5}{@{}l}{\textit{Stage~1 $\rightarrow$ Bench2Drive v0.0.4}}\\
AutoMoT & 84.3 & \textbf{95.1} & 66.4 & \textbf{88.6}\\
SimLingo & 86.6 & \textbf{92.7} & 70.5 & \textbf{83.2}\\
\midrule
\multicolumn{5}{@{}l}{\textit{Stage~2 $\rightarrow$ Fail2Drive}}\\
AutoMoT & \fTwoAutoMoTDS & \textbf{\fTwoStageTwoAutoMoTDS} & \fTwoAutoMoTAverageSR & \textbf{\fTwoStageTwoAutoMoTSR}\\
SimLingo & 77.2 & \textbf{80.9} & 67.2 & \textbf{80.0}\\
\bottomrule
\end{tabularx}
\end{table}

\begin{figure}[t]
\centering
\includegraphics[width=\linewidth]{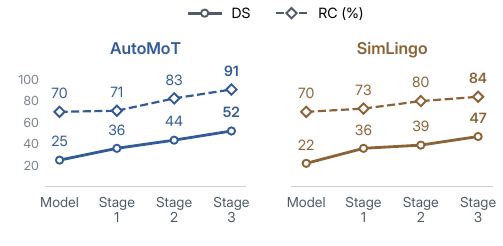}
\caption{\textbf{Frozen-stage transfer to Longest6.} Both driving models remain fixed across Stages~1--3. Solid lines show DS and dashed lines show RC (\%). Longest6 is excluded from system revision.}
\label{fig:longest6-evolution}
\end{figure}

On Longest6, AutoMoT progresses from 25 DS alone to 36, 44, and 52 across Stages~1--3, while SimLingo progresses from 22 to 36, 39, and 47 (Fig.~\ref{fig:longest6-evolution} and Appendix~E). Each system retains its sensor configuration and driving model, perception, and VLM checkpoints across these stages. SimLingo's progression from 36 to 39 to 47 demonstrates accumulated skill and orchestration capability with the driving model and shared perception fixed. Final RC reaches 91\% and 84\%, respectively, from 70\% for each model alone.

\subsection{Execution Architecture Comparisons}
\label{sec:component-comparisons}
We compare the final Stage~3 system on 62 Bench2Drive v0.0.3 routes, comprising one route per interaction type and 18 additional high-interaction routes where AutoMoT performs poorly. AutoMoT alone scores 83.72 DS on these routes (Table~\ref{tab:orchestration-ablation}). Appendix~F details the variants.
\begin{table}[t]
\centering
\caption{\textbf{Execution architecture comparisons} on 62 Bench2Drive v0.0.3 routes, with one run per configuration. Fixed-schedule updates cover model inference, perception, and orchestration.}
\label{tab:orchestration-ablation}
\small\setlength{\tabcolsep}{3pt}
\begin{tabularx}{\linewidth}{@{}Xrr@{}}
\toprule Configuration & DS $\uparrow$ & SR (\%) $\uparrow$\\
\midrule
\textbf{Full \method} & \textbf{99.77} & \textbf{96.77}\\
\midrule
AutoMoT alone & 83.72 & 62.90\\
AutoMoT + perception \& safety & 86.97 & 66.13\\
\midrule
Decision-tree task selection & 92.11 & 80.65\\
Rule-based model replacement & 71.83 & 43.55\\
Whole-vehicle skill takeover & 84.47 & 66.13\\
Model-controlled cruising/following & 95.68 & 88.71\\
Direct handover & 95.20 & 88.71\\
Fixed-schedule updates & 94.75 & 85.48\\
\bottomrule
\end{tabularx}
\end{table}

The full system succeeds on 60 routes, versus 39 for the model and 41 with perception and safety alone. Task execution expands the range of interactions the system can complete beyond safety filtering. Replacing VLM judgments with PDM-Lite-trained decision trees reduces successes to 50, showing the contribution of the VLM to task selection.

Rule-based model replacement uses bounded best-first search over kinematic motion primitives, guided by perceived drivable space and navigation. With the same perception, skills, and controllers, it yields 71.83 DS and 27 successes. This comparison shows the contribution of learned motion generation within the shared execution architecture.

Whole-vehicle skill takeover tests capability composition by assigning all three channels to an engaged skill. The same rule-based planner supplies missing motion references, and the model resumes outside takeover. This yields 84.47 DS and 41 successes, versus 99.77 DS and 60 successes when model and skill outputs are combined across channels. On the 18 high-interaction routes, success rises from 9 to 18.

Model-controlled cruising/following uses model inference in place of the local cruising and following controller and yields 55 successes. Direct handover removes handover readiness checks and the monitored model trial, also succeeding on 55 routes. The variants incur five and four vehicle collisions, respectively, versus zero for the full system.

Figure~\ref{fig:cases} illustrates how tasks guide execution. The obstacle-passage task completes lane return after its VLM proposal disappears. During traffic interaction, model steering remains combined with skill speed until motion and traffic conditions support handover. The task retains responsibility through the model trial until observed continuation is accepted.

Fixed-schedule updates query the model on three of every ten control steps and update perception and orchestration on a fixed 67\% of steps. Skill and controller implementations remain unchanged, with feedback and safety checks at every step. This variant reaches 94.75 DS and 85.48\% SR. The full system combines per-step state updates with task-dependent model calls, gaining 5.02 DS and seven successful routes over this fixed schedule.

\subsection{Adaptive Scheduling and Agent Latency}
\begin{table}[t]
\centering
\caption{\textbf{Agent latency on RTX 5090.} Times in ms include VLM inference. SimLingo uses chain-of-thought reasoning.}
\label{tab:latency}
\small\setlength{\tabcolsep}{3pt}
\begin{tabularx}{\linewidth}{@{}Xrrr@{}}
\toprule Method & Median $\downarrow$ & Mean $\downarrow$ & p90 $\downarrow$\\
\midrule
SimLingo & \latencySimLingoBaseMedian & \latencySimLingoBaseMean & \latencySimLingoBasePNinety\\
\quad\textbf{+ \method} & \textbf{\latencySimLingoHarnessMedian} & \textbf{\latencySimLingoHarnessMean} & \latencySimLingoHarnessPNinety\\
\midrule
AutoMoT & \latencyBaseMedian & \latencyBaseMean & \latencyBasePNinety\\
\quad\textbf{+ \method} & \textbf{\latencyHarnessMedian} & \latencyHarnessMean & \latencyHarnessPNinety\\
\bottomrule
\end{tabularx}
\end{table}

We measure agent latency, including VLM inference, by replaying preprocessed Longest6 inputs on an RTX 5090 outside CARLA with common Longest6 execution-path weights (Appendix~F).

\method lowers SimLingo's median latency by 80.2\% and mean from \latencySimLingoBaseMean{} to \latencySimLingoHarnessMean{}\,ms (Table~\ref{tab:latency}). AutoMoT's median falls from \latencyBaseMedian{} to \latencyHarnessMedian{}\,ms, while its mean rises from \latencyBaseMean{} to \latencyHarnessMean{}\,ms. The additional cost of periodic VLM updates contributes to the higher p90.

\begin{table}[t]
\centering
\caption{\textbf{Model inference utilization on Longest6.} Percentage of control steps over complete runs.}
\label{tab:control-allocation}
\small\setlength{\tabcolsep}{3pt}
\begin{tabularx}{\linewidth}{@{}Xrr@{}}
\toprule Model inference & AutoMoT & SimLingo\\
\midrule
\textbf{Model inference skipped} & \textbf{\longestPolicySkipShare} & \textbf{\longestSimLingoSkipShare}\\
New model inference & \longestPolicyCallShare & \longestSimLingoCallShare\\
\bottomrule
\end{tabularx}
\end{table}

Across complete Longest6 runs, model inference is skipped on \longestPolicySkipShare\% of steps with AutoMoT and \longestSimLingoSkipShare\% with SimLingo (Table~\ref{tab:control-allocation}). Task updates and feedback control continue using current observations, while valid motion references and skill outputs allow execution to proceed between model calls.

\FloatBarrier
\paragraph{Limitations.}
Evaluation covers four models in simulation. Periodic VLM updates produce a latency tail. Real-world deployment requires further evaluation.

\section{Conclusion}
\method coordinates frozen driving models, skills, and tools through shared physical state and persistent tasks. Joint orchestration of execution and inference improves closed-loop driving. A stable task interface supports capability accumulation, demonstrated by frozen-stage transfer across benchmarks and driving models.

\label{maintextend}
\section*{LLM Usage Statement}
Codex (gpt-6-sol) and Claude Opus 5.5 assisted system implementation, analysis of execution records, code revision, and running evaluations. GPT-6 reviewed VLM supervision labels. Claude and GPT also assisted with manuscript editing.

\label{referencesstart}
{\small\bibliographystyle{ieeenat_fullname}\bibliography{references}}
\clearpage
\appendix
\setcounter{table}{0}
\setcounter{figure}{0}
\setcounter{equation}{0}
\renewcommand{\thetable}{S\arabic{table}}
\renewcommand{\thefigure}{S\arabic{figure}}
\renewcommand{\theequation}{S\arabic{equation}}
\renewcommand{\theHtable}{appendix.\arabic{table}}
\renewcommand{\theHfigure}{appendix.\arabic{figure}}
\renewcommand{\theHequation}{appendix.\arabic{equation}}
\raggedbottom
\twocolumn[\begin{center}\Large\bfseries Supplementary Material\end{center}\vspace{0.5em}]
This supplement provides implementation details (Appendix~A), shared physical state and VLM outputs (B), skill contracts (C), handover conditions (D), system evolution (E), execution analysis and measurement protocols (F), and benchmark results and baseline sources (G).
\section{Implementation and VLM Inputs}
Model adapters preserve each driving model's sensor, coordinate, navigation, and output conventions.

\subsection{Perception Reuse and Model Inputs}
AutoMoT, TransFuser++, and the evaluated BridgeDrive all use TransFuser-family perception with object detection and BEV-semantic outputs. AutoMoT's native model uses frozen TransFuser++ camera--LiDAR features. The perception harness applies detection and BEV-semantic heads from the same pretrained checkpoint to these features. This checkpoint also belongs to the TransFuser++ model's pretrained ensemble, and its adapter uses the same camera and LiDAR configuration. Tracking and occupancy processing turn predictions and point clouds into shared physical state.

BridgeDrive uses the LEAD adaptation with a TFv6 TransFuser backbone, a CenterNet detection head, and a BEV-semantic decoder~\cite{bridgedrive,lead}. Its native inputs include cameras, LiDAR, and radar. The adapter preserves these inputs and the frozen TFv6 model. The common PDM-Lite-trained TransFuser++ module supplies shared physical state to the harness through detection and BEV-semantic outputs. All detectors retain their pretrained weights.

SimLingo retains its visual encoder and camera-only model inputs, while its complete system additionally uses the common camera--LiDAR module. For each model, sensor configuration and learned checkpoints remain fixed across Stages~1--3, so comparisons between these stages evaluate system revisions with the same learned perception.

\subsection{VLM Inputs and Update Schedule}

The runtime uses three update schedules. A local VLM query is launched in the background every five control steps. Reference validity, control demand, and handover checks determine when to call the driving model. The orchestrator updates tasks and channel allocation at every control step, using new VLM proposals when available. Proposals retain their observation timestamps, allowing the interface to reject stale proposals while continuing admitted tasks from current shared physical state. Each control step advances 0.05\,s of simulation time. The distilled VLM supplies the online proposals.

The VLM is Qwen3-VL-2B~\cite{qwen3vl} adapted with LoRA~\cite{lora}. Its inputs combine front-view images at simulation-time offsets of $-1.5,-1.0,-0.5,0$ seconds, a compact description of shared physical state, and perception-derived relation candidates. The orchestrator maintains active-task state separately. An empty proposal requests no new task, while an active task continues according to its execution conditions.

The LoRA adapter uses rank 64, scaling factor 128, and dropout 0.05 on the language model's attention and feed-forward projections. The four front-view frames are resized to $512\times256$ pixels.

Appendix~F describes the wall-clock latency measurement protocol.

\subsection{VLM Training}
The VLM is trained with cross-entropy averaged over answer tokens, with prompt tokens masked from the loss. Relation and auxiliary visual-question-answering examples receive weights of 1.0 and 0.5, respectively.

We first adapt Qwen3-VL-2B to driving using PDM-Lite contexts~\cite{hiddendata} and SimLingo training question-answer pairs, then train its LoRA adapter on relation labels. Relation training samples from a dataset of 116,903 examples, including 4,000 auxiliary question-answer pairs from the SimLingo training split. Training runs for 1,200 optimizer updates with an effective batch size of 16. AdamW uses learning rate $5\times10^{-5}$, zero weight decay, a 60-step linear warmup, and cosine decay. Open-loop training is completed during initial system construction, and the VLM remains fixed across all reported stages.

\subsection{Teacher-Reviewed Relations}
\label{sec:s-exposure}
VLM training labels are constructed from expert driving clips and yielding episodes. Rules use annotated future scene evolution to propose candidate relations, and GPT-6 reviews every candidate. During initial system construction, development records of scenes requiring skill intervention inform the teacher prompts. The teacher receives scene images, a description of shared physical state, and numbered candidates. It can confirm or reject a candidate, correct its scene bindings, or mark the available observations as insufficient. Confirmed candidates and candidates with corrected bindings become training labels. Future scene annotations are used only to construct offline labels.

Each target relation identifies its family, subject, counterpart, region, and task.

\subsection{Control Between Model Calls}
At each control step, the runtime tracks the selected references or forwards native commands on their assigned channels. Cached references expire after ten control steps and can be invalidated earlier by scene, task, or navigation changes. Reuse during cruising and following requires current perception and compatible task state. A required hold includes an observed red-light wait or an admitted skill's waiting command. A navigation change, departure after a traffic-rule stop, invalid motion reference, or safety-filter request triggers new model inference. Pending handover can also require fresh model output for the readiness check, with ownership still governed by the skill's handover conditions. Tracking errors request new model inference when the current reference fails validity checks. With a valid reference, controller selection can instead address sustained tracking error (Appendix~C). These conditions implement the main paper's inference rule.

For reference tracking, the system projects the rear axle onto the supplied path and checks lateral and heading error, path continuity, and remaining extent. Let $v$ be current speed, $t_r$ the response allowance, $b$ the braking deceleration, and $d_0$ the clearance margin. The required stopping distance is
\begin{equation}
 d_{\rm support}=vt_r+\frac{v^2}{2b}+d_0.
\end{equation}
Both the observed corridor and the remaining path must provide this distance, and the path must extend through the tracking lookahead. Failing any of these checks requests a fresh model plan. Following additionally bounds target speed using the lead vehicle's observed motion and available stopping distance. Reference reuse therefore depends on current geometry and traffic as well as reference age.

\section{Shared Physical State and VLM Outputs}
\subsection{Perception and Tracked Scene State}
Frozen TransFuser++ perception processes onboard observations. Task state and shared physical state guide additional checks of ambiguous detections. Temporal tracking and LiDAR geometry supply motion and observed surfaces, including LiDAR tracks beyond the spatial extent of the TransFuser++ BEV window. These observations are stored with object identities, geometry, timestamps, and validity, so VLM proposals, skill conditions, and continuation checks refer to the same interaction.

\subsection{State Representation}
Each element of shared physical state $\E_t$ is stored as a record
\begin{equation}
 e=(\nu,\mathfrak t,\sigma,t_{\rm obs},t_{\rm exp},\mathfrak f,p,\mathcal J),
\end{equation}
where $\nu$ is the value, $\mathfrak t$ its type, $\sigma$ its validity state, $t_{\rm obs}$ the observation time, $t_{\rm exp}$ its expiry information, $\mathfrak f$ the coordinate frame, $p$ the producer, and $\mathcal J$ links to the supporting observations. Modules assess freshness from these observations and their timestamps. A recently delivered estimate may still describe an old observation. Table~\ref{tab:s-status} defines the record validity states.
\begin{table}[t]
\centering
\caption{\textbf{Physical-state record validity.} Record validity determines whether an observation supports a task or transition. Task conditions additionally check the required physical constraints.}
\label{tab:s-status}
\small
\setlength{\tabcolsep}{3pt}
\begin{tabularx}{\linewidth}{@{}l>{\raggedright\arraybackslash}X@{}}
\toprule
Record validity & Interpretation\\
\midrule
\texttt{OK} & Available and usable under the producer's validity conditions.\\
\texttt{UNKNOWN} & The observation does not establish the required fact.\\
\texttt{STALE} & The supporting observation is no longer current.\\
\texttt{INCONSISTENT} & Required observations or scene bindings conflict.\\
\texttt{UNSUPPORTED} & The requested quantity or relation is outside the producer's supported interface.\\
\bottomrule
\end{tabularx}
\end{table}

Moving-object records combine geometry and motion estimates. Traffic streams group tracked vehicles by their observed paths and motion, recording spacing and membership as individual vehicles enter or leave the interaction. Static obstruction surfaces are represented separately from moving road users. Unknown occupancy remains distinct from observed clearance.

Conflict regions and traversal corridors use observed geometry and the original task route. Temporary navigation guidance changes the model input while preserving the corridor that defines the task. Coordinates are transformed using the corresponding observation frame and time. Admission and release require resolved coordinates and sufficiently current evidence.

\subsection{Relation Vocabulary}
VLM proposals describe driving tasks through scene relations. Each relation specifies a family, subject, counterpart, region, and requested task. The counterpart is the interacting object or traffic stream. Entity bindings resolve against the shared physical state. The five relation families and their execution requirements are defined in the main paper's Skill Library subsection. Each family describes a driving requirement independently of the skill implementation.

Table~\ref{tab:specialists} details the four skill families and representative implementations. Emergency yielding and traffic-rule obligations share rule compliance. Different compatible implementations can fulfill a relation, and one skill can serve several related tasks. A proposal can instantiate a task once its scene bindings resolve to observations and the skill applicability and channel-allocation checks pass.

\section{Physical Execution Architecture and Skill Contracts}
\begin{table}[t]
\centering
\caption{\textbf{Skill families and representative implementations.} Channels follow the main paper's notation. Each task requests the channels required by its maneuver. The arrow denotes a change during execution.}
\label{tab:specialists}
\small\setlength{\tabcolsep}{3pt}
\begin{tabularx}{\linewidth}{@{}>{\raggedright\arraybackslash}p{.24\linewidth}>{\centering\arraybackslash}p{.20\linewidth}>{\raggedright\arraybackslash}X@{}}
\toprule Skill family & Channels & Implementation\\
\midrule
Traffic interaction & $\Lat,\Lon$ & Dynamic programming for departure and speed, with feedback execution\\
Obstacle passage & $\Lon\to$ all three & Quadratic programming for passage paths, space--time planning, and lane return\\
Rule compliance & $\Lon$ or all three & Signal/STOP handling uses $\Lon$. Emergency yielding can use a reference and lane change on all three channels.\\
Route guidance & $\Nav$ & Adjusted navigation inputs guide model-generated motion\\
\bottomrule
\end{tabularx}
\end{table}

\subsection{Tasks and Skill Instances}
Skill families address recurring geometric, temporal, and traffic-rule requirements. Each skill implementation computes the task's constraints from measured geometry and motion. Persistent tasks link these requirements to scene entities and retain execution status and channel responsibility. Admission grants the skill instance access to the required channels. Pending requests wait until applicability and channel conditions are met. Once admitted, a task continues across VLM updates according to its completion and handover conditions.

Binding first resolves the counterpart and region named by a relation, then identifies the observations needed by the skill. For a merge, these include the front and rear vehicles bracketing the gap. Task identity can remain stable while these observations change. Before the vehicle commits to a maneuver, new valid observations can update the binding. After commitment, the task evaluates changes as part of the continuing maneuver or recovery.

\subsection{Skill Contract}
The skill contract defined in the main paper specifies applicability $\Omega_i$, required channels $W_i$, output generation $\kappa_i$, completion $d_i$, and handover readiness $h_i$. Each applied channel has a single source.

Rule-compliance tasks retain obligations and progress across observations, including traffic-signal and STOP handling and emergency yielding. Channel ownership governs skill contributions, while the safety filter checks the composed motion command from any skill family or the model.

\paragraph{Binding and admission.}
For a proposed relation $r_t\in\mathcal R_t$, the interface binds its scene entities to current observations:
\begin{equation}
b_i=\bind(r_t,\E_t)\in\mathcal B\cup\{\bot\},
\end{equation}
where $\mathcal B$ is the set of supported physical bindings and $\bot$ indicates an unresolved binding. Admission requires valid bindings, applicable skill conditions, and available channels, following the main paper's admission rule. The channel set is reserved together, preserving the skill's coupled control requirements. A request whose physical conditions are met may still wait for admission while another task owns the required channels.

\subsection{Model Adapter and Controller Selection}
The adapter constructs navigation inputs and converts model outputs to the shared execution interface. Trajectory-producing models use a tracker, while control-producing models retain native commands. Route guidance adjusts navigation before a model query. Lateral references are selected before tracking, and longitudinal contributions adjust the command on their allocated channel. The safety filter checks the composed command. Execution monitoring can request a new prediction when continuation requires a fresh reference.

The orchestrator selects between feedback and learned controllers. The feedback controller uses PID tracking. The learned controller is a frozen reference-tracking multilayer perceptron (MLP) with three 512-unit hidden layers, LayerNorm, and SiLU activations. It receives a 20-point path, target speed, ego-motion feedback, the previous command, and disturbance estimates. Its nine predicted longitudinal/steering command pairs use a tanh output, with the first pair applied. Training minimizes closed-loop tracking error through a calibrated vehicle-dynamics model. Feedback control tracks model or skill references, while learned control is selected for eligible model-reference tracking.

\paragraph{Controller selection from execution feedback.}
The tracking-error monitor evaluates the most recent historical world-frame reference aged between 0.25 and 0.6\,s. This history measures tracking response, while execution uses references that pass current validity checks. Using an earlier reference exposes lateral error that reanchoring a new path at the current position would hide. Projection onto this reference gives lateral error $e_y$ and heading error $e_\psi$. Its target speed gives speed error $e_v$. The normalized response error is
\begin{equation}
\epsilon_t=\max\!\left\{
 m_t\frac{|e_y|}{0.5\,\mathrm m},\;
 m_t\frac{|e_\psi|}{15^\circ},\;
 q_t\frac{|e_v|}{2.5\,\mathrm{m/s}}
\right\}.
\end{equation}
Here $m_t$ enables geometric errors at speeds of at least 0.5\,m/s. The speed gate $q_t$ is active when the target changes by at most 1.25\,m/s and the absolute speed error has not improved by more than 0.1\,m/s since the reference was applied. This distinguishes a changing motion request from an inadequate tracking response.

Selecting the learned controller requires a valid model reference, an available tracker, and model-reference tracking in control of both motion channels. An error above one for 0.3\,s switches tracking to the learned controller. Error below 0.5 for 0.5\,s returns tracking to the feedback controller. A 3\,s limit on learned controller use and 1\,s cooldown bound repeated switching. Four alternations between acceleration and braking commands within 2\,s (active pedal at least 0.5, opposite pedal at most 0.05) provide an additional trigger, using the tracking commands actually applied at positive target speed. A conflicting task or invalid reference resets the monitor. These rules change the controller while retaining the selected motion reference.

\subsection{Traffic-Interaction Planning and Execution}
The traffic-interaction skill plans motion along an available path while the task maintains its interaction goal. The path can come from the model or a task-specific reference. Tracked vehicles are propagated along their observed courses using estimated speed. Oriented body overlap with ego path samples produces a space--time occupancy grid over an 8\,s horizon at 0.1\,s intervals. Crossing and same-direction traffic contribute conflict and headway constraints, respectively. Crossings and merges share the planner while current geometry and motion specify its constraints. Static occupancy limits the space available for stopping and passage.

Dynamic programming searches longitudinal position, speed, actuator-response state, and the preceding acceleration command. Each transition integrates a candidate acceleration and checks occupancy, speed bounds, and task-specific motion constraints. Terminal states either clear the conflict or retain sufficient stopping space before it. The objective accounts for completion time, clearance margin, acceleration effort, and changes in acceleration. State discretization and dominance pruning retain competing feasible trajectories. An unsuccessful coarse search is retried at finer position and speed resolution.

Execution uses the first command and retains the remaining plan with its original time reference. Before reuse, the remaining commands are rolled forward from the current measured state against refreshed occupancy and the same motion constraints. A valid retained plan can therefore continue between searches. When no feasible departure exists, an uncommitted task waits or decelerates. An already committed task instead compares recovery trajectories using collision and motion-constraint violations before comfort costs. The task remains active during recovery, and the common safety filter checks the applied command.

\subsection{Obstacle-Passage Planning and Execution}
An obstacle-passage relation identifies the obstructing entity and the intended passage side. The skill projects measured obstacle extent, free road boundaries, and tracked motion into route coordinates $(s,l)$. Static occupancy bounds the admissible lateral interval $[l_k^-,l_k^+]$ for the ego body at each longitudinal sample. The basic passage path solves
\begin{equation}
\begin{aligned}
\min_{\boldsymbol l}\quad &\sum_k \bigl[(l_k-l_k^{\rm ref})^2
 +2(Dl_k)^2+100(D^2l_k)^2\bigr]\\
&+20(Dl_0-\tan\psi_0)^2,\\
\text{s.t.}\quad &l_k^-\leq l_k\leq l_k^+ ,
\end{aligned}
\end{equation}
where $l_k^{\rm ref}$ is the reference lateral offset, $D$ is a finite difference along the route, and $\psi_0$ is the initial heading relative to it. The initial lateral position is fixed. The planner samples the corridor at 1\,m intervals over a 60\,m horizon and truncates it where the bounds become infeasible. A curvature-constrained formulation additionally carries forward the current path's initial geometry.

Tracked traffic supplies space--time occupancy along the passage. Speed search selects among candidate acceleration profiles, checking body overlap and the ability to stop before an infeasible corridor. A viable reference is tracked in closed loop. Lane return remains part of the task after passing the obstacle. Measured obstacle extent sets clearance bounds, road geometry sets available passage space, and tracked traffic constrains timing in same-direction or oncoming lanes. The same implementation can therefore handle different obstruction types and layouts within its applicability conditions. The VLM identifies the interaction, while the skill computes its geometry and timing.

\subsection{Safety Filtering}
The common safety filter checks the composed command against applicable traffic-rule obligations and observed contact constraints. Signal and STOP tasks maintain their obligations across steps. Contact checks use tracked bodies and LiDAR occupancy along the proposed motion. Braking or speed restrictions can override longitudinal contributions from the model or any skill. The task remains responsible during these adjustments, with release governed by completion and handover checks.

\subsection{Relation to Other Runtime Architectures}
Table~\ref{tab:s-runtime-comparison} compares what related harnesses coordinate during execution. In \method, channel ownership retains task responsibility while compatible model and skill outputs contribute together. Task state and execution feedback guide controller selection, reference reuse, and model inference.
\begin{table*}[t]
\centering
\caption{\textbf{Runtime coordination mechanisms.} The comparison summarizes the mechanisms described by each work. AutoAgent0 is concurrent work.}
\label{tab:s-runtime-comparison}
\small\setlength{\tabcolsep}{4pt}
\begin{tabularx}{\textwidth}{@{}l>{\raggedright\arraybackslash}p{.23\textwidth}X@{}}
\toprule
System & What is coordinated & Mechanism\\
\midrule
AutoAgent0~\cite{autoagentzero} & Candidate motion trajectory & Verify nominal proposals against refined scene context. Generate and revise recovery proposals when needed.\\
Harness VLA~\cite{harnessvla} & Analytic or VLA primitive call & Compose analytic staging and transport with a frozen VLA for contact-rich execution.\\
RoboHarness~\cite{roboharness} & Heterogeneous policy for a subtask & Use execution memory for capability-aware routing and prepare a compatible state for the next policy.\\
\method & Task-owned channels and model calls & Combine model and skill contributions during a maneuver. Retain task state through handover and coordinate controllers, reference reuse, and inference.\\
\bottomrule
\end{tabularx}
\end{table*}

\section{Task Continuity and Handover}
\begin{figure*}[t]
\centering
\includegraphics[width=\textwidth]{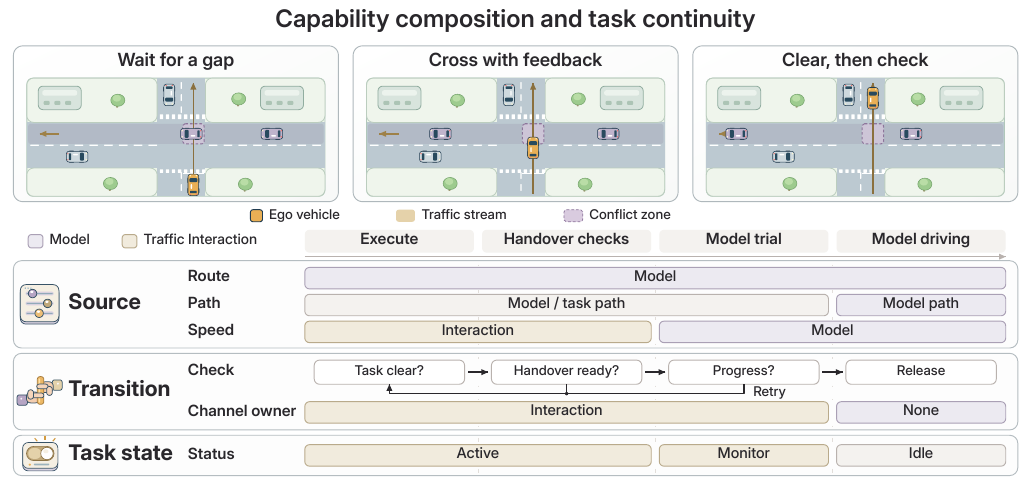}
\caption{\textbf{Capability composition and task continuity in one traffic interaction.} Model and skill outputs are combined across channels. A monitored trial applies model outputs while the task retains channel ownership. Route, Path, and Speed correspond to navigation, lateral, and longitudinal channels. Task status is Active during execution and handover checks, Monitor during the trial, and Idle after task termination. Gray-violet denotes the model and ochre the interaction task.}
\label{fig:orchestration}\label{fig:lifecycle}
\end{figure*}
\subsection{Lifecycle}
Task status follows the same three phases shown in Fig.~\ref{fig:orchestration}. \textbf{Active} covers waiting, execution, and handover checks. \textbf{Monitor} applies model outputs during a monitored model trial, and \textbf{Idle} follows task termination. Returning a channel can precede task termination. A request awaiting admission is not yet active. Table~\ref{tab:s-transitions} lists task operations from nomination to release, including binding updates and recovery. Output-source changes and responsibility release are separate state transitions.
\begin{table*}[t]
\centering
\caption{\textbf{Task lifecycle operations.} The applicable operations depend on the task, its execution state, and current observations. A task can return individual channels before it ends.}
\label{tab:s-transitions}
\small
\setlength{\tabcolsep}{5pt}
\renewcommand{\arraystretch}{1.08}
\begin{tabularx}{\textwidth}{@{}p{.15\textwidth}*{3}{>{\raggedright\arraybackslash}X}@{}}
\toprule
Operation & Required conditions & Task update & Effect on execution\\
\midrule
Nominate & Valid VLM proposal & Create or update a task request & Await admission before acquiring channels.\\
Admit & Valid binding, skill applicability, available channels & Assign the requested channels to the task & Permit outputs on those channels.\\
Wait & An admitted task awaiting a feasible opportunity & Maintain the task and its assigned channels & Continue observation and waiting control.\\
Commit & The executor starts following a feasible maneuver plan & Mark the maneuver as committed & Retain responsibility for continuation and recovery.\\
Rebind & A reviewed, valid binding for the same task before commitment & Update the continuing task's valid bindings & Preserve task identity through the update.\\
Complete & The task meets its fulfillment conditions & Mark the requirement as fulfilled & Check whether the model is ready to take over.\\
Recover & Execution review calls for recovery after unsuitable continuation or handover & Resume skill control or recovery planning on held channels & Retain task responsibility.\\
Model trial & For a monitored handover, completion and readiness checks pass and model output is valid & Retain channel ownership during the trial & Apply model control and observe the resulting motion.\\
Release & A validated plan needs fewer channels or the task ends & Remove the task as owner of the returned channels & Make those channels available for reassignment.\\
\bottomrule
\end{tabularx}
\end{table*}

A traffic-interaction task can hold the speed channel while waiting for a gap. It can update its binding before commitment. Once it begins the maneuver, it remains responsible for continuation or recovery as traffic changes.

\subsection{Completion and Handover Readiness}
Completion evaluates local task fulfillment. Handover readiness evaluates whether the model can continue from the pose, velocity, route context, and observed conflicts resulting from the intervention. For example, after the vehicle passes an obstruction, the task can remain active while it guides the vehicle back into its lane.

When handover includes a model trial, the model supplies control while the task retains channel ownership. The runtime releases those channels once it verifies that the model can continue. If the trial fails, the task can resume control. Channels can also be returned while a task is still active. For example, after lane return, an obstacle-passage task can give back its channels and continue monitoring the model path. A new conflict can prompt it to request the channels again.

\subsection{Traffic-Interaction Handover Conditions}
The traffic-interaction skill evaluates task clearance, conflict continuity, front and rear spacing, and the receiving model's operating conditions. The conflict check projects braking, constant-speed, and acceleration continuations over a 2\,s horizon against current occupancy. Front and rear checks compare observed spacing with response-based distance requirements. When the receiving model has a speed cap, the handover check compares it with the current speed.

Table~\ref{tab:s-handover} summarizes the checks. The longitudinal spacing check uses the response-distance model of RSS~\cite{rss}. For rear and front speeds $v_r,v_f\geq0$, the required bumper gap is
\begin{equation}
\delta(v_r,v_f)=\left[v_r\rho+\frac{a\rho^2}{2}
+\frac{(v_r+a\rho)^2}{2b_r}-\frac{v_f^2}{2b_f}\right]_+,
\end{equation}
where $[z]_+=\max(z,0)$. The implementation uses response time $\rho=0.6$\,s, acceleration bound $a=2$\,m/s$^2$, and braking magnitudes $b_r=4$ and $b_f=6$\,m/s$^2$. The front check evaluates $\delta(v_{\rm ego},v_{\rm lead})$, while the rear check reverses the roles to $\delta(v_{\rm follower},v_{\rm ego})$. Conflict projections span 2\,s at 0.1\,s intervals using current occupancy. These checks assess readiness for a trial. The trial then measures model continuation.

\begin{table}[t]
\centering
\caption{\textbf{Traffic-interaction handover checks.} These checks support entry into a monitored model trial.}
\label{tab:s-handover}
\small
\begin{tabularx}{\linewidth}{@{}lX@{}}
\toprule
Check & Physical evidence\\
\midrule
Task clearance ($d_i$) & Ego body has cleared the task's conflict region.\\
Conflict continuity & Current occupancy supports the projected handover continuations.\\
Front spacing & Gap to the observed vehicle ahead satisfies the response-distance requirement.\\
Rear spacing & Associated followers have sufficient observed separation at handover.\\
Model speed cap & Current speed is within the receiving model's speed cap, when configured.\\
Valid output & The receiving model supplies valid output for the required channels.\\
\bottomrule
\end{tabularx}
\end{table}

Before generating new outputs, the orchestrator reviews the model commands applied during the trial and the resulting motion. The task retains channel ownership during the model trial, which lasts up to 2\,s. The orchestrator accepts handover once the model makes forward progress or an observed object ahead justifies waiting.

\subsection{Channel Allocation}
Channel allocation is non-preemptive. Existing holders retain their channels. Requests acquire free channels only when their bindings and applicability conditions are satisfied (Appendix~C). Channel-set expansion is atomic, preserving the previous allocation if a conflict occurs. A successful contraction releases the removed channels. A task's outputs can be applied only to channels it holds.

\section{System Evolution}
\label{sec:s-evolution}
System evolution uses execution records to improve skill execution and orchestration offline. Revisions change applicability conditions, output generation, and computation scheduling through the same task interface. All stages retain the five relation families, four skill families, and three control channels. The driving model, perception networks, and VLM keep the same weights throughout these stages.

\subsection{Offline Revision and Inheritance}
Crossings and merges share a traffic-interaction planner. Each task supplies its own geometry and traffic motion, so a change to the planner can serve both maneuvers. Later stages inherit the revised planner through the same skill contract. For either maneuver, the orchestrator can pair the planner's speed output with a compatible model path.

Execution records link observations and task bindings to the outputs applied on each channel and the resulting motion. Coding agents use these records to locate an unmet requirement, revise the skill or its coordination rules, and evaluate the revised behavior.

\subsection{Physical Mechanisms Behind Revisions}
\paragraph{Yielding along the intended path.}
During a turn, a pedestrian near the exit can conflict with the intended path while lying outside the vehicle's current heading direction. Projecting the interaction along the intended path lets yielding account for the space the vehicle is about to traverse. The projection is recomputed as road orientation and participant positions change. Stage~2 inherits this mechanism from Stage~1.

\paragraph{Obstacle passage before crossing.}
A vehicle partially obstructing the ego lane before a straight junction creates a dependency between passage and crossing. Admitting the crossing task first can reserve channels needed to pass the obstruction. The revised applicability condition defers crossing when measured clearance permits passage to finish before the junction, allowing the passage task to acquire the channels first.

The condition requires an active same-direction maneuver, a fresh measured obstruction endpoint, and straight-ahead junction commands. It projects the obstruction's far endpoint toward the junction and checks that the remaining separation accommodates the ego half-length and a junction clearance. Applicability follows this measured spatial relationship, allowing the same rule to handle different obstacle positions and junction layouts.

\subsection{System Stages}
Stage~1 establishes the initial library and runtime on Bench2Drive v0.0.3. Stage~2 extends the inherited system on v0.0.4 and also includes closed-loop runs on CARLA training routes. Stage~3 refines the system on Fail2Drive (Table~\ref{tab:s-stage-changes}).

\begin{table}[t]
\centering
\caption{\textbf{Capability accumulation across system stages.} Later stages inherit the preceding implementation with model, perception, and VLM weights fixed.}
\label{tab:s-stage-changes}
\small
\begin{tabularx}{\linewidth}{@{}lX@{}}
\toprule Stage & System changes\\
\midrule
1 & Initial integration of traffic interaction, obstacle passage, rule compliance, and route guidance through task binding, channel ownership, and handover conditions.\\
2 & Refined traffic observations and route guidance, static-occupancy protection, and updated applicability conditions.\\
3 & Local cruising and following control, expanded reference reuse, and refined model-inference conditions.\\
\bottomrule
\end{tabularx}
\end{table}

Benchmark evaluations freeze the corresponding stage and cover both development and held-out sets under the main-paper evaluation protocol. Stage~1 transfers to v0.0.4 and Stage~2 to Fail2Drive before revision on the respective target benchmark. Longest6 is reserved for evaluating frozen systems throughout development.

\subsection{Development and Held-Out Sets}
\label{sec:s-route-groups}
Development sets contain routes used for system revision and regression checks, while held-out sets contain routes excluded from these development activities. The development and held-out sets contain 82/138 routes on Bench2Drive v0.0.3, 60/160 on v0.0.4, and 68/132 on Fail2Drive. Table~\ref{tab:s-route-groups} compares AutoMoT with and without \method on the same routes within each set. The Bench2Drive results use the same route evaluations as the main-paper comparisons. The Fail2Drive sets pool Base and Gen routes.

\begin{table*}[t]
\centering
\caption{\textbf{Performance on development and held-out sets.} Paired comparisons of AutoMoT and AutoMoT with \method. Gains are computed before rounding. SR is a percentage.}
\label{tab:s-route-groups}
\small\setlength{\tabcolsep}{5pt}
\begin{tabular}{@{}llrrrrrr@{}}
\toprule
& & & \multicolumn{3}{c}{DS $\uparrow$} & \multicolumn{2}{c}{SR (\%) $\uparrow$}\\
\cmidrule(lr){4-6}\cmidrule(l){7-8}
Benchmark & Set & Routes & Model & + \method & $\Delta$ & Model & + \method\\
\midrule
Bench2Drive v0.0.3 & Development & 82 & 76.31 & 97.37 & +21.07 & 45.12 & 91.46\\
 & Held-out & 138 & 97.63 & 97.91 & +0.29 & 93.48 & 93.48\\
\midrule
Bench2Drive v0.0.4 & Development & 60 & 76.76 & 95.62 & +18.86 & 48.33 & 88.33\\
 & Held-out & 160 & 87.16 & 98.43 & +11.27 & 73.13 & 95.00\\
\midrule
Fail2Drive & Development & 68 & 65.63 & 78.99 & +13.36 & 54.41 & 77.94\\
 & Held-out & 132 & 79.46 & 91.75 & +12.29 & 68.94 & 94.70\\
\bottomrule
\end{tabular}
\end{table*}

On v0.0.3, development-set DS improves by 21.07 and SR rises from 45.12\% to 91.46\%. On held-out routes, AutoMoT alone already reaches 97.63 DS, and \method maintains 97.91 with SR unchanged at 93.48\%. Held-out DS gains of 11.27 on v0.0.4 and 12.29 on Fail2Drive show that the shared capabilities also improve performance beyond the development sets.

\subsection{Frozen-Stage Transfer to Longest6}
The transfer protocol evaluates AutoMoT and SimLingo alone and with the same system stages frozen before Longest6 evaluation. Each configuration uses the 36 routes of Longest6 v2 and the Leaderboard~2 evaluator. Within Stages~1--3, each model retains the same sensor configuration, driving-model checkpoint, and shared perception checkpoint. SimLingo additionally tests reuse of stages constructed with AutoMoT. Evaluation uses the official 200\,m global-plan spacing and minimum-speed penalty.

Table~\ref{tab:stage-transfer} reports route-averaged DS and RC. The initial gain includes system construction, while subsequent gains measure the transfer of accumulated revisions under fixed model and perception weights.

\begin{table}[t]
\centering
\caption{\textbf{Frozen-stage transfer to Longest6.} Each model and the VLM remain fixed across stages. Both models use the same system stages. RC is a percentage. Values are rounded to integers.}
\label{tab:stage-transfer}
\small\setlength{\tabcolsep}{4pt}
\begin{tabularx}{\linewidth}{@{}Xrrrr@{}}
\toprule
& \multicolumn{2}{c}{AutoMoT} & \multicolumn{2}{c}{SimLingo}\\
\cmidrule(lr){2-3}\cmidrule(l){4-5}
Configuration & DS $\uparrow$ & RC $\uparrow$ & DS $\uparrow$ & RC $\uparrow$\\
\midrule
Model alone & \longestBaselineDS & \longestBaselineRC & \longestSimLingoBaselineDS & \longestSimLingoBaselineRC\\
Stage 1 & \longestStageOneDS & \longestStageOneRC & \longestSimLingoStageOneDS & \longestSimLingoStageOneRC\\
Stage 2 & \longestStageTwoDS & \longestStageTwoRC & \longestSimLingoStageTwoDS & \longestSimLingoStageTwoRC\\
Stage 3 & \textbf{\longestStageThreeDS} & \textbf{\longestStageThreeRC} & \textbf{\longestSimLingoStageThreeDS} & \textbf{\longestSimLingoStageThreeRC}\\
\bottomrule
\end{tabularx}
\end{table}

\section{Execution Analysis and Measurement Protocol}
\subsection{Task Continuity and Handover}
The main paper's qualitative cases show how task continuity follows physical progress. During lane borrowing, the VLM stops proposing obstacle passage while a 2.63\,m lateral offset remains. Retaining the task lets lane return continue. During traffic interaction, the task combines model steering with skill-generated speed until handover readiness checks pass. Table~\ref{tab:s-case-events} relates these decisions to the maneuver stages.

\begin{table*}[t]
\centering
\caption{\textbf{Task continuity through handover.} Model output application and task release are separate events.}
\label{tab:s-case-events}
\small
\begin{tabularx}{\textwidth}{@{}lrXX@{}}
\toprule
Case & Time (s) & Relation or execution evidence & System behavior\\
\midrule
Lane borrowing & 16.05 & Obstacle-passage relation & Wait for opposing traffic.\\
 & 25.05 & Obstacle-passage relation & Borrow the adjacent lane and pass the blockage.\\
 & 27.05 & No new proposal, lateral offset 2.63\,m & Keep lane-return control active.\\
 & 31.05 & Vehicle aligned with original lane & Apply model motion.\\
\midrule
Traffic interaction & 11.05 & Traffic-interaction relation & Wait for a suitable gap.\\
 & 13.05 & Traffic-interaction relation & Join the traffic stream.\\
 & 17.05 & No new proposal, model outputs full braking & Defer handover while continuing the task.\\
 & 17.45 & Handover checks pass & Start a model trial.\\
 & 17.65 & Model trial accepted & Release the task.\\
 & 18.05 & Traffic-rule relation & Apply model control with the rule check.\\
\bottomrule
\end{tabularx}
\end{table*}

In the traffic-interaction case, the vehicle exceeds the receiving model's configured speed cap (10.91 versus 9.72\,m/s), and the model outputs full braking. The task continues driving until the handover checks pass, then monitors model motion before releasing task responsibility.

\subsection{Capability Composition}
The separate merge episode in the main paper's capability-composition table shows how capability composition retains useful model output during skill execution. Applied steering follows the model while the skill supplies throttle in place of the model's full-brake command. The task holds both motion channels, retaining recovery responsibility while using the model's steering. After release, the controller tracks a model reference generated 0.10\,s earlier without new model inference. Model outputs can therefore contribute through capability composition, handover, and reference reuse.

\subsection{Adaptive Scheduling and Inference Events}
Inference utilization is computed over the complete Longest6 runs of the final Stage~3 system with AutoMoT and SimLingo. Each control step is assigned to one execution path using model-inference events and the applied steering, throttle, and brake commands. Model calls follow the main paper's inference rule. Required holds, valid model references, and executable skill plans satisfy $\eta_t^{\rm hold}$, $\eta_t^{\rm reuse}$, and $\eta_t^{\rm plan}$, respectively. Feedback control executes these outputs, adjusting cruising and following speed from current observations. The categories distinguish vehicle motion, reference reuse, feedback control, and traffic-interaction planning. Initialization and warmup are excluded. Table~\ref{tab:s-control-breakdown} reports the distribution.

\begin{table*}[t]
\centering
\caption{\textbf{Execution-path distribution over complete Longest6 runs.} Rounded percentages of control steps assigned to mutually exclusive runtime paths. Feedback control can also operate within other paths. Bold rows are group totals.}
\label{tab:s-control-breakdown}
\small\setlength{\tabcolsep}{5pt}
\begin{tabularx}{\textwidth}{@{}Xrr@{}}
\toprule Execution path & AutoMoT (\%) & SimLingo (\%)\\
\midrule
\textbf{Model inference skipped} & \textbf{71.5} & \textbf{71.3}\\
\quad Stopping hold & 19.3 & 21.8\\
\quad Feedback control while moving & 18.0 & 16.2\\
\quad Cached model reference & 11.7 & 10.1\\
\quad Feedback control while stationary & 10.2 & 11.5\\
\quad Traffic-interaction execution between planning calls & 4.8 & 4.9\\
\quad Other feedback-control updates & 4.5 & 3.6\\
\quad Red-light handling & 3.0 & 3.2\\
\midrule
\textbf{New model inference} & \textbf{28.5} & \textbf{28.7}\\
\quad Model inference while moving & 15.9 & 14.1\\
\quad Model inference while stationary & 10.6 & 12.5\\
\quad Model inference during traffic-interaction planning & 2.0 & 2.0\\
\bottomrule
\end{tabularx}
\end{table*}

The table groups applied execution paths, which can share the same inference-skipping condition. Task responsibility is tracked separately and can span categories. A skill can use a model-generated path, and a model trial can apply the model's command while its physical task remains active. The system skips model inference on 71.5\% of steps with AutoMoT and 71.3\% with SimLingo. Cached model references account for 11.7\% and 10.1\% of steps, respectively. The remaining skipped steps execute required holds, valid motion references, or skill plans through feedback control.

\subsection{Agent Computation Latency}
We replay recorded Longest6 sensor inputs in temporal order on an RTX 5090 outside CARLA, executing the complete agent and measuring wall-clock latency at each control step. System latency weights the measured execution-path distributions by their frequencies in the 36 AutoMoT Longest6 routes, with equal weight per route. Both models use these weights. Timing spans preprocessed sensor inputs to final control commands, including system preparation, perception, VLM and driving-model inference, orchestration, planning, tracking, the safety filter, and synchronization. Initialization, warmup, and visualization are excluded.

\subsection{Execution Architecture Comparisons}
Table~\ref{tab:s-ablation-subsets} separates the comparison of the final system on Bench2Drive v0.0.3 into one route per interaction type and 18 additional high-interaction routes where AutoMoT performs poorly. Each configuration is evaluated once per route for the paired analysis below. Direct handover uses the full system's skills and output generators but applies model control and immediately releases responsibility at task completion. In both configurations, ongoing tasks persist when the VLM supplies no new proposal. The comparison isolates the full system's handover readiness checks and monitored model trial, which retain responsibility until model continuation is accepted.

The model-controlled cruising/following variant uses model inference in place of local cruising and following control, retaining skill execution and the safety filter.

Decision-tree task selection replaces the VLM's judgments with shallow decision trees fitted only to the PDM-Lite training set. The rules operate on perception-derived candidates and supply proposals to the same execution architecture, retaining persistent tasks, capability composition, handover checks, and inference conditions.

AutoMoT + perception and safety retains auxiliary perception and the safety filter while disabling task admission and skill execution.
\begin{table*}[t]
\centering
\caption{\textbf{Stage~3 execution architecture comparisons by route group.} Driving score (DS) and success rate (SR) on 44 routes covering all Bench2Drive interaction types and 18 additional high-interaction routes.}
\label{tab:s-ablation-subsets}
\small\setlength{\tabcolsep}{6pt}
\begin{tabularx}{\textwidth}{@{}XXrr@{}}
\toprule Route group & Configuration & DS $\uparrow$ & SR (\%) $\uparrow$\\
\midrule
44 interaction types & \textbf{Full \method} & \textbf{99.67} & \textbf{95.45}\\
 & AutoMoT alone & 87.78 & 72.73\\
 & AutoMoT + perception \& safety & 90.15 & 72.73\\
 & Decision-tree task selection & 94.79 & 84.09\\
 & Rule-based model replacement & 73.82 & 47.73\\
 & Whole-vehicle skill takeover & 89.18 & 72.73\\
 & Model-controlled cruising/following & 97.17 & 90.91\\
 & Direct handover & 97.72 & 93.18\\
 & Fixed-schedule updates & 95.60 & 86.36\\
\midrule
18 high-interaction routes & \textbf{Full \method} & \textbf{100.00} & \textbf{100.00}\\
 & AutoMoT alone & 73.81 & 38.89\\
 & AutoMoT + perception \& safety & 79.19 & 50.00\\
 & Decision-tree task selection & 85.55 & 72.22\\
 & Rule-based model replacement & 66.95 & 33.33\\
 & Whole-vehicle skill takeover & 72.93 & 50.00\\
 & Model-controlled cruising/following & 92.05 & 83.33\\
 & Direct handover & 89.05 & 77.78\\
 & Fixed-schedule updates & 92.66 & 83.33\\
\bottomrule
\end{tabularx}
\end{table*}

\paragraph{Rule-based model replacement.}
This comparison replaces the driving model with receding-horizon kinematic motion planning. A bounded best-first search expands constant-curvature motion primitives from measured ego motion, with pose and curvature defining each search state. Curvature and its change along the path are bounded. Candidate ranking combines path length, curvature, navigation progress, lane alignment, road clearance, and reference continuity. Inputs comprise perception-derived semantic BEV, detected boxes, ego motion, and the current and next navigation targets (TP/NTP). Vehicle-footprint checks test drivable space and occupancy. Path curvature and free stopping distance bound speed, while traffic-interaction skills handle temporal conflicts. The planner supplies path and speed references through the decision harness. Perception, the VLM, skills, reference reuse, controllers, and safety filtering are retained.

\paragraph{Whole-vehicle skill takeover.}
An engaged skill receives exclusive navigation, lateral, and longitudinal channels. Skills retain their native outputs. The same motion-primitive planner supplies any missing path or speed reference from perceived geometry and navigation targets. During takeover, navigation adjustments condition this planner, and model trajectories do not supply the skill's missing outputs. The learned model remains available for driving outside skill execution and for handover. Perception, the VLM, feedback controllers, and safety filtering are retained. This comparison tests complete takeover against capability composition, where a model path can remain useful while a skill regulates speed.

\paragraph{Fixed-schedule updates.}
This variant retains the model, perception and VLM weights, skills, and controller implementations. It queries the model on three of every ten control steps and updates perception and orchestration on a fixed 67\% of steps. Skill applicability, planning, channel allocation, and controller-selection rules operate at these updates. Between updates, selected controllers track retained references using current vehicle state. The VLM keeps its original cadence. Sensor ingestion, feedback control, and safety checks continue at every control step.

\paragraph{Paired route analysis.}
We pair outcomes by route and bootstrap the 62 route pairs with 20,000 resamples. The 95\% percentile intervals for the full system's DS advantage are [1.35, 8.52] over direct handover, [1.13, 7.74] over model-controlled cruising/following, and [8.06, 17.86] over perception and safety alone. These intervals describe variation across the evaluated routes.

The full system incurs zero vehicle collisions, compared with four across three routes for direct handover and five across five routes for model-controlled cruising/following. Perception and safety alone incur 17 vehicle collisions across 16 routes.

\FloatBarrier
\section{Benchmark Results and Baseline Sources}
\label{sec:s-public-references}
On Bench2Drive v0.0.3, the AutoMoT comparison uses our paired route evaluations. SimLingo and TransFuser++ model-alone scores follow the Bench2Drive leaderboard~\cite{b2dexpertrefs}. On v0.0.4, SimLingo model-alone and LEAD expert scores follow the benchmark release~\cite{b2d004}, while AutoMoT and TransFuser++ model-alone scores use our evaluations. All \method results come from our evaluations. Privileged experts in the main paper use ground-truth scene state~\cite{b2dexpertrefs,lead}.

\paragraph{Fail2Drive.}
Table~\ref{tab:f2d-full} reports the split-wise and average DS, SR, and HM for each model alone and with \method. AutoMoT-alone results come from our evaluations, while published SimLingo and BridgeDrive model-alone results follow~\cite{autoagentzero}. HM is computed within each split, and average HM is the mean of the two split HMs.
\begin{table*}[!t]
\centering
\caption{\textbf{Model and system performance on Fail2Drive.} \method rows describe the accumulated Stage~3 system. In-distribution (Base) and generalization (Gen) each contain 100 routes. HM is the harmonic mean of DS and SR within each split. Average takes the mean of the two splits for each metric. SR is a percentage.}
\label{tab:f2d-full}
\small\setlength{\tabcolsep}{4pt}
\begin{tabularx}{\textwidth}{@{}Xrrrrrrrrr@{}}
\toprule
 & \multicolumn{3}{c}{Base} & \multicolumn{3}{c}{Gen} & \multicolumn{3}{c}{Average}\\
\cmidrule(lr){2-4}\cmidrule(lr){5-7}\cmidrule(lr){8-10}
Method & DS $\uparrow$ & SR $\uparrow$ & HM $\uparrow$ & DS $\uparrow$ & SR $\uparrow$ & HM $\uparrow$ & DS $\uparrow$ & SR $\uparrow$ & HM $\uparrow$\\
\midrule
BridgeDrive & 91.6 & 95.0 & 93.3 & 81.9 & 75.0 & 78.3 & 86.8 & 85.0 & 85.8\\
\quad\textbf{+ \method} & \fTwoBridgeHarnessBaseDS & \fTwoBridgeHarnessBaseSR & \fTwoBridgeHarnessBaseHM & \fTwoBridgeHarnessGenDS & \fTwoBridgeHarnessGenSR & \fTwoBridgeHarnessGenHM & \fTwoBridgeHarnessAverageDS & \fTwoBridgeHarnessAverageSR & \fTwoBridgeHarnessAverageHM\\
\midrule
SimLingo & 82.6 & 79.3 & 80.9 & 71.7 & 55.0 & 62.2 & 77.2 & 67.2 & 71.6\\
\quad\textbf{+ \method} & \fTwoSimLingoBaseDS & \fTwoSimLingoBaseSR & \fTwoSimLingoBaseHM & \fTwoSimLingoGenDS & \fTwoSimLingoGenSR & \fTwoSimLingoGenHM & \fTwoSimLingoDS & \fTwoSimLingoAverageSR & \fTwoSimLingoAverageHM\\
\midrule
AutoMoT & \runTenIDDS & \runTenIDSR & \runTenIDHM & \runTenOODDS & \runTenOODSR & \runTenOODHM & \fTwoAutoMoTDS & \fTwoAutoMoTAverageSR & \fTwoAutoMoTAverageHM\\
\quad\textbf{+ \method} & \runNineIDDS & \runNineIDSR & \runNineIDHM & \runNineOODDS & \runNineOODSR & \runNineOODHM & \fTwoAutoMoTHarnessDS & \fTwoAutoMoTHarnessAverageSR & \fTwoAutoMoTHarnessAverageHM\\
\bottomrule
\end{tabularx}
\end{table*}

\FloatBarrier
\end{document}